\documentclass[letterpaper]{article}

\usepackage{natbib,alifeconf}  
\usepackage{url,hyperref}
\usepackage{booktabs}
\usepackage{svg}
\usepackage{subcaption}
\usepackage{caption}
\usepackage{hyperref}
\usepackage{amsmath}
\usepackage{amsfonts}
\usepackage{float}
\usepackage[export]{adjustbox}
\usepackage{cleveref}

\title{ALIFE2025 template}

\title{EngramNCA: a Neural Cellular Automaton Model of Memory Transfer}

\author{
    Etienne Guichard$^{1}$,
    Felix Reimers$^{1}$, \and
    Mia Kvalsund$^{2}$, \and
    Mikkel Lepper$\o$d$^{3}$, \and
    Stefano Nichele$^{1,4}$\\
    \mbox{}\\
    $^1$$\O$stfold University College, Halden, Norway \\
    $^2$University of Oslo, Oslo, Norway \\
    $^2$Simula Research Laboratory, Oslo, Norway \\
    $^2$Oslo Metropolitan University, Oslo, Norway \\
    stefano.nichele@hiof.no
} 

\begin{document}

\maketitle

\begin{abstract}
This study introduces EngramNCA, a neural cellular automaton (NCA) that integrates both publicly visible states and private, cell-internal memory channels, drawing inspiration from emerging biological evidence suggesting that memory storage extends beyond synaptic modifications to include intracellular mechanisms. The proposed model comprises two components: GeneCA, an NCA trained to develop distinct morphologies from seed cells containing immutable "gene" encodings, and GenePropCA, an auxiliary NCA that modulates the private "genetic" memory of cells without altering their visible states. This architecture enables the encoding and propagation of complex morphologies through the interaction of visible and private channels, facilitating the growth of diverse structures from a shared "genetic" substrate. EngramNCA supports the emergence of hierarchical and coexisting morphologies, offering insights into decentralized memory storage and transfer in artificial systems. These findings have potential implications for the development of adaptive, self-organizing systems and may contribute to the broader understanding of memory mechanisms in both biological and synthetic contexts.
\end{abstract}



\textbf{Data/Code}: A web version of this article with videos is available \href{https://etimush.github.io/EngramNCA/}{here}, while the Github repository is available \href{https://github.com/etimush/EngramNCA}{here} and the code is available on Colab \href{https://colab.research.google.com/github/etimush/MemoryNCA/blob/main/EngramNCA.ipynb}{here}. Images that represent videos are hyperlinked to their respective video in the web version.

\section{Introduction}

The currently most accepted theory of memory in biological and artificial brains is the synapses dogma 
\citep{mayford2012synapses}, that is, the place where information is stored is at synapses. Therefore, it is believed that the storage of memories happens through changes in synaptic strengths 
\citep{cajal1894fine}. This view is challenged by emerging evidence that memory is also present privately within neural cells (and potentially other types of cells) \citep{abraham2019plasticity}, and synapses are merely a visible reverberation of such private memories 
\citep{gold2021central}. Compelling examples are the evidence of memory transfer via injection of RNA extracted from trained aplysia into untrained animals 
\citep{bedecarrats2018}, the memory transfer via a soup of trained planaria fed to untrained organisms 
\citep{mcconnell1967modern}, and the decapitation of planaria that retained long-term memories after head regeneration 
\citep{shomrat2013automated}. Such findings suggest that RNA transfer between cells can influence cellular functions and behaviors, including memory storage and transmission. Practically, cells can transfer RNA molecules to their surroundings via extracellular vesicles (membrane-bound particles) that encapsulate RNA and other molecules, or by tunneling nanotubes that form a direct physical connection between cells. 

In this work, we propose a Neural Cellular Automaton (NCA) 
\citep{mordvintsev2020growing}
 model that allows for private memory storage within cells and memory transfer across cells. By replicating such memory mechanisms, we might equally be able to enable richer intercellular communication and more reliable generation of complex patterns, compared to classical NCAs that rely exclusively on publicly visible cell states. We name this new model EngramNCA, with reference to \textit{engram} which in neuropsychology indicates the physical substrate of memory in biological organisms or, in other words, the means by which memories are stored. 
 
 We extend the NCA model in two ways:
\begin{itemize}
  \item Each cell has a publicly visible state (comprised of a certain number of public CA channels) that can be accessed by neighboring cells and used as available information for computing cell state updates. This is akin to information transmitted through synapses to neighboring cells.
  \item Each cell has a private memory (comprised of a certain number of private CA channels) that is visible only to the cell itself. This represents some sort of internal state stored more permanently in a cell, that can be possibly transmitted or made available to other cells in a manner akin to RNA transfer (or other plausible molecular transfer mechanisms).
\end{itemize}

When CA cells update their state, their internal neural network (a neural network with identical weights for all cells) acts upon the visible channels in the neighborhood, in addition to the cell’s own private memory. The result is the interplay of the visible component and the (private) memory stored within the cell. As such, a focus is to train the NCA to utilize its cells internal memory to full capacity, since it is not directly accessible by other cells. Therefore, it has to learn to produce a reverberation of it through its visible channels in order to carry out the collective computation in coordination with the neighbors.

This coordination is achieved by first training an NCA to grow from a seed cell to a set of primitive morphologies. The only cell initially containing the genetic information (private memory) representing the target morphology is the seed cell itself. Each morphology is encoded by a different private representation. Therefore, this NCA has to learn to achieve the global task only through its visible channels. The neural network weights of this first NCA are then frozen and do not change. We name this first neural network GeneCA.

Subsequently, another NCA is trained in addition to the previous, which does not manipulate and change the visible information of cells, but can only change their private genetic memory, i.e., a set of private channels. As such, it has to learn a mechanism to regulate the private information such that the previously learned primitive morphologies (or their combination) can be utilized and activated at the right time and in the right place. We name this second neural network GenePropCA, short for GenePropagationCA.

The GenePropCA mechanism is an abstraction for RNA molecular transmission between cells, which provides a form of functionality transfer. The transmitted functionality is stored in the cell’s internal memory and, in turn, affects the behavior of the GeneCA (the cell’s visible state).

Note that in biological systems, the two types of information transfer, i.e., synaptic transfer and RNA transfer, may happen at two different timescales. Typically, spikes through synapses are transmitted faster than RNA transfers. However, in our framework we have tested with different timescales (basically running the GeneCA faster than the GenePropCA, or vice versa) without affecting the functionality. Overall, the system learns to adjust the rate of updates from one of the two NCA to solve the task at hand.

We test the proposed NCAs on different morphogenetic tasks, including the growth of simple yet stable primitives, out-of-distribution mixtures of primitives, more complex morphologies obtained by the combination of basic primitives, and moving primitives such as Lenia gliders.

\section{Related Works}
\subsection{On Neural Cellular Automata}
Neural cellular automata (NCAs) merge the classical framework of cellular automata with modern deep learning by replacing hand-crafted update rules with neural networks that are optimized via gradient-based methods. Earlier studies in cellular automata (e.g., \citep{Wolfram2002}) established that simple local rules can lead to complex emergent behavior. Building on these ideas, \citep{mordvintsev2020growing} introduced the “Growing Neural Cellular Automata” model, which trains a convolutional network with gradient-based learning to iteratively update a grid so that, starting from a single seed, complex images emerge and can regenerate after damage.

Seminal work by \citep{miller2004evolving} explored evolving self-repairing, self-regulating cellular automata to produce the “French flag” pattern, demonstrating how evolutionary algorithms can discover diverse rule sets for complex morphogenesis. \citep{nichele2017neat} extended morphogenetic CAs with neuro-evolution of neural networks as CA rules. \citep{pontescritical} showed a NCA evolved to criticality. This was further studied by \citep{Guichard2024}. \citep{kvalsund2024sensor} discussed NCA a model for a distributed neocortex.

A range of extensions have broadened the capabilities of NCAs. \citep{randazzo2020self} explored self-classification tasks. \citep{hernandez2021neural} presented NCA Manifold, creating embedding spaces of different morphogenetic processes. \citep{tesfaldet2022attention} developed an attention-based NCA that integrates self-attention mechanisms into the update process, allowing each cell to dynamically focus on relevant features in its local neighborhood and capture longer-range dependencies. \citep{grasso2022empowered} proposed Empowered Neural Cellular Automata, incorporating an information-theoretic objective (empowerment) to promote robust, coordinated behaviors. \citep{palm2022variational}
proposed Variational Neural Cellular Automata, a probabilistic generative model that captures diverse emergent dynamics through variational inference. \citep{mordvintsev2022growing} proposed an isotropic version of NCAs, while \citep{pande2023hierarchical}
presented Hierarchical NCA to model emergent behaviors at different scales.

Extensions in application domains further illustrate the versatility of the NCA framework. \citep{variengien2021towards}
applied NCAs to evolve controllers for a cart-pole system. \citep{horibe2022severe}
demonstrated the use of NCAs for regenerating soft robot morphologies, showcasing their potential for self-repair akin to biological regeneration. \citep{najarro2022hypernca} studied a hypernetwork approach, named HyperNCA, that grows artificial neural networks with a developmental process. More recently, \citep{Reimers2023} introduced a pathfinding NCA with local self-attention, training automata to collectively explore environments and locate energy sources. \citep{randazzo2024simulating} exploited an NCA as an evolvable plant biome. Very recently, Differentiable Logic Cellular Automata by \citep{Pietro-Miotti-Eyvind-Niklasson-Ettore-Randazzo-Alexander-Mordvintsev2025-jb}
integrates differentiable logic gates with traditional neural cellular automata to achieve discrete, interpretable update rules while maintaining the benefits of gradient-based training.

\subsection{Related to Memory Engrams}

One work that greatly inspired us is \citep{Pappadopoulos2023}, where the incorporation of additional channels to encode for different morphologies is discussed. Additionally, \citep{sudhakaran2022goal} proposes a goal-encoder which serves a similar function to our proposed GenePropCA, while \citep{bessonov2015model} shares a similar inspiration to our work and uses a cellular automaton as test-bench. Also, \citep{winge2023artificial} shares a rather similar inspiration, however they use neurons with cell memory and their execution is rather different. More broadly, we were greatly inspired by \citep{MitchellCheney2024} and \citep{HartlLevin2024}, arguing that living systems encode robust, adaptable generative models that guide development and regeneration. Such works emphasize that biological lineages accumulate “memory engrams”—latent representations that enable organisms to reliably rebuild form and function despite perturbations. This concept is central in our idea of EngramNCA, where the model leverages dual-channel memory (public and private states) to simulate the distributed storage and transfer of memory observed in biological systems.

\section{Methods}

In this section, we introduce the changes to the overall architecture from \citep{mordvintsev2020growing}, and detail the implementation of the two new models (GeneCA and GenePropCA, respectively), that form the backbone of the EngramNCA model, including illustrations of their specific architecture and how they interact with each other.

\begin{figure}[t!]
\centering
\includegraphics[width=0.47\textwidth]{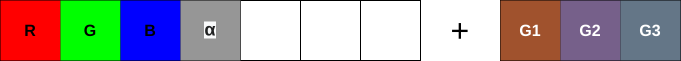}
\caption{NCA augmentation through gene embedding channels, where the channels of the cell are represented, including the RGB channels, the alpha channel, three additional channels, plus three gene channels (G1,G2, and G3) representing the private tape.}
\label{fig:embeddings}
\end{figure}

Taking inspiration from RNA memory transfer observed Planarians and Aplysia, we augment the NCA hidden state with a "private tape". Instead of extending the classic NCA model by appending the tape to the hidden states, we choose to sacrifice hidden states by privatizing them, this ensures that any improvements seen are not due to an increase in model size (please see Appendix 1 for masking of hidden channels to investigate task deterioration with less available channels). The hypothesis here is that the private tape augmentation can be exploited to encode different modalities of behaviour, such as growing different shapes or representing different computations. An illustration is provided in Figure \ref{fig:embeddings}, where the channels of the cell are represented, including the RGB channels, the alpha channel, three additional channels, plus three gene channels (G1, G2, and G3) representing the private tape.

\begin{figure}[t!]
\centering
\includegraphics[width=0.47\textwidth]{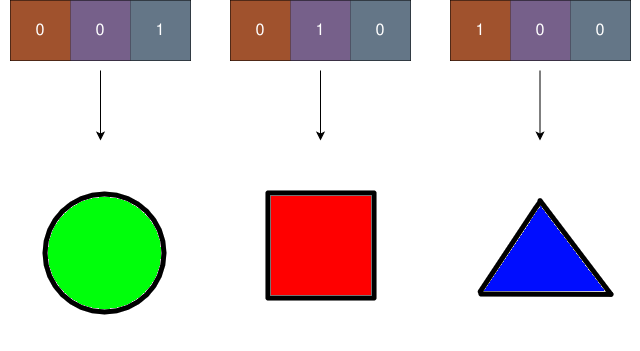}
\caption{Binary embedding of geometric morphologies and their respective encoding.}
\label{fig:shapes}
\end{figure}

\begin{figure}[t!]
\centering
\includegraphics[width=0.47\textwidth]{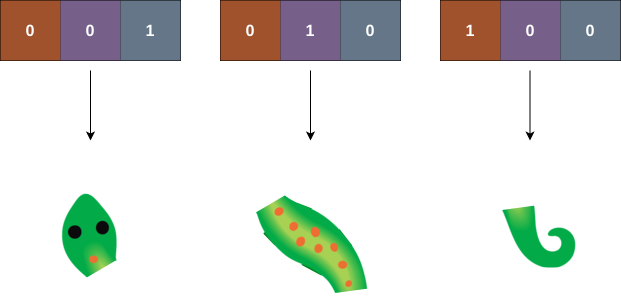}
\caption{Binary embedding of lizard body sections and their respective encodings.}
\label{fig:lizzParts}
\end{figure}

We chose to encode different "primitives" in the tape through binary representation. These primitives can consist of simple shapes such as convex polygons or more complex, convex, compound shapes such as body segments of a lizard emoji. Examples of encodings are represented in Figure \ref{fig:shapes} and Figure \ref{fig:lizzParts}, where three basic geometric morphologies and lizard parts are given, respectively.

The tape length has to be chosen such that $2^{gene\_channels}$ is greater or equal to the number of primitives. This ensures that, due to the binary representation, there are enough distinct binary encodings for each primitive. However, due to reasons that will be discussed later, it is often preferable to have a buffer of genes that are not used in the binary encoding. There is a delicate balance between the number of gene bits and shared hidden channels that needs to be reached as privatisation of channels incurs a performance penalty (see Appendix 1 for details).

\subsection{GeneCA}

\begin{figure*}[t!]
\centering
\includegraphics[width=0.94\textwidth]{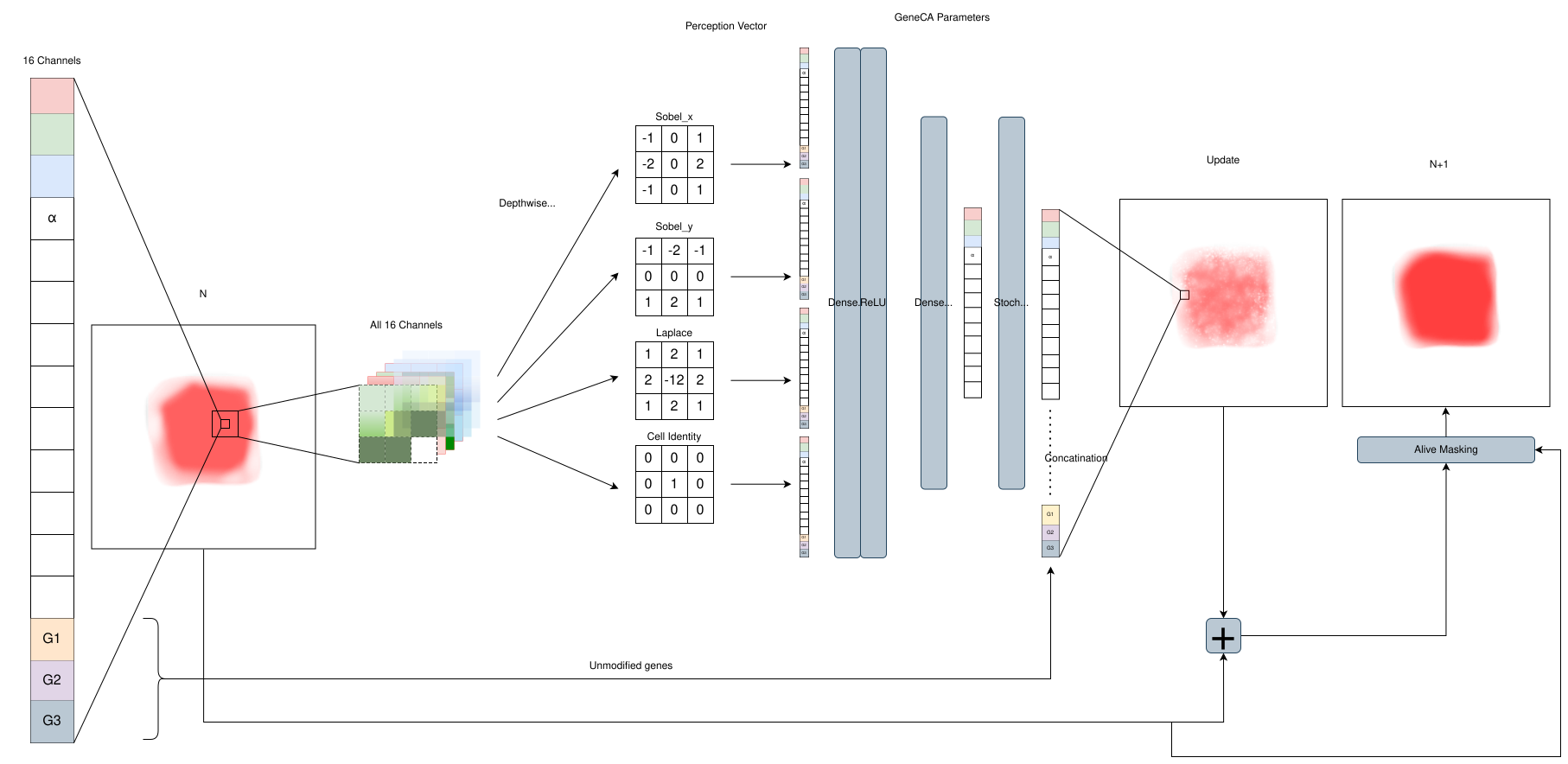}
\caption{One step of the GeneCA}
\label{fig:GeneCAStep}
\end{figure*}

The first model (Figure \ref{fig:GeneCAStep}), which we name GeneCA, is tasked with learning the primitive embedding. It takes inspiration from the \citep{mordvintsev2020growing} and is extended to include a laplacian convolution filter as found in \citep{mordvintsev2022growing}. Notably, the parameters only produce a partial cell-state update-vector that comprises all non-gene channels. After the stochastic update step, the gene channels are cloned and appended to the update vector to prevent degradation over time. This leads to no change or propagation of gene channel information throughout the NCA cell. It is worth noting that since the only living cell that is initialized at the beginning of each run is a single seed cell, all cells besides the seed cell have no information in their gene channels (i.e., all zeros). This means that the GeneCA learns to grow the primitive morphologies from a seed cell without relying on a mechanism to transmit the gene channel information to other cells. Additionally, this signifies that cells do not practically observe the private tape of neighboring cells.

\subsection{GenePropCA}

\begin{figure*}[t!]
\centering
\includegraphics[width=0.94\textwidth]{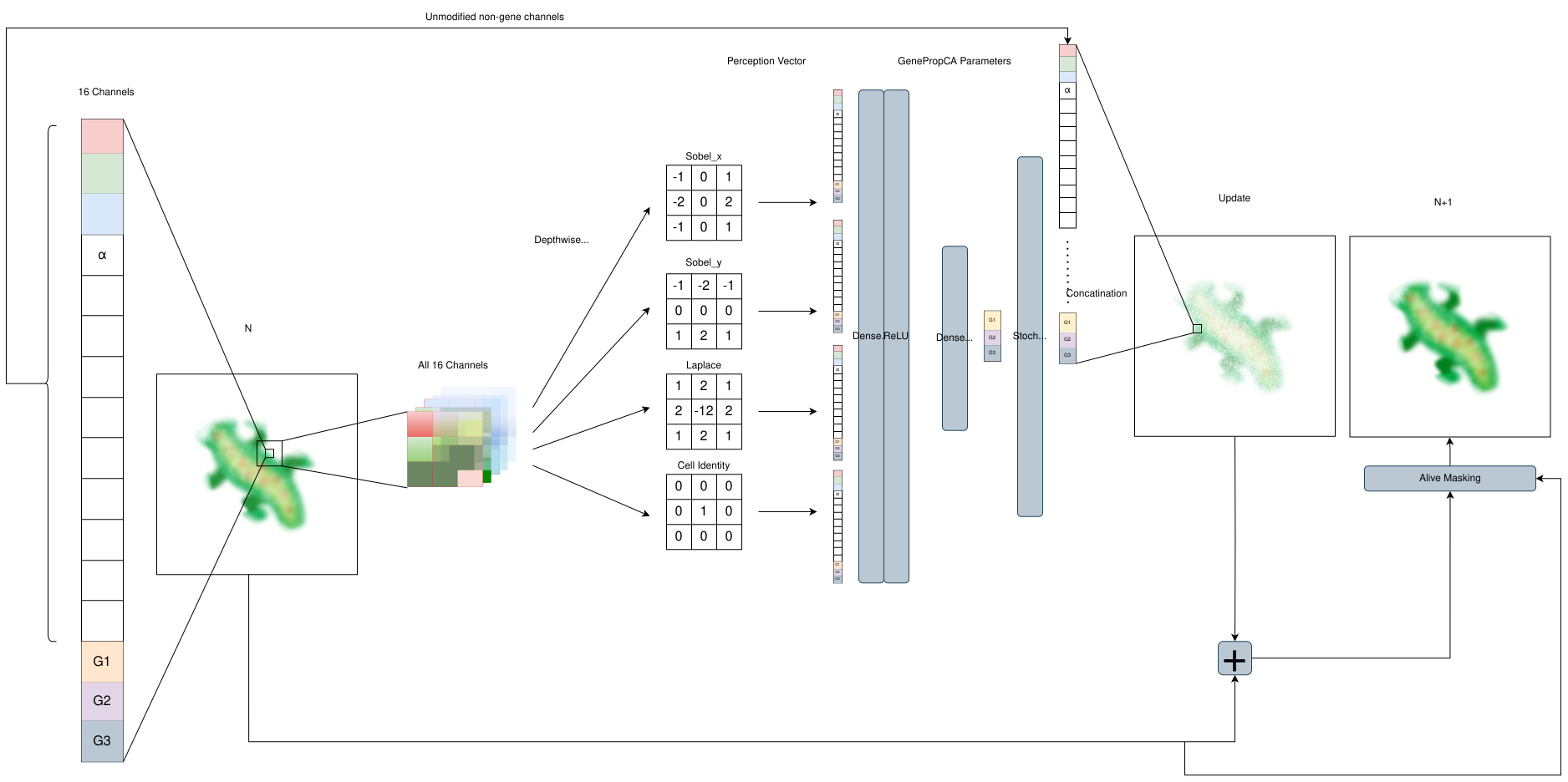}
\caption{One step of the GenePropCA}
\label{fig:GenePropCaStep}
\end{figure*}

The GenePropCA model (Figure \ref{fig:GenePropCaStep}) is tasked with exploiting the gene embedding learned by the GeneCA. By conditioning the primitive growth in the GeneCA to a specific embedding, these embeddings now carry latent information about the primitives. We assume that the latent information in each embedding can be modified to produce out-of-distribution behaviour, thus allowing the GenePropCA to both propagate and modify the genes to achieve a desired goal. The architecture of the GenePropCA is similar to that of the GeneCA, where the neural network parameters produce a partial cell state update. In this case, only the genes channels are updated. Like in the GeneCA, the unmodified cell states are cloned and appended after the asynchronous partial update, leading to the GenePropCA having no influence over non-gene channels.

\subsection{EngramNCA - An Ensemble Model}

\begin{figure*}[t!]
\centering
\includegraphics[width=0.99\textwidth]{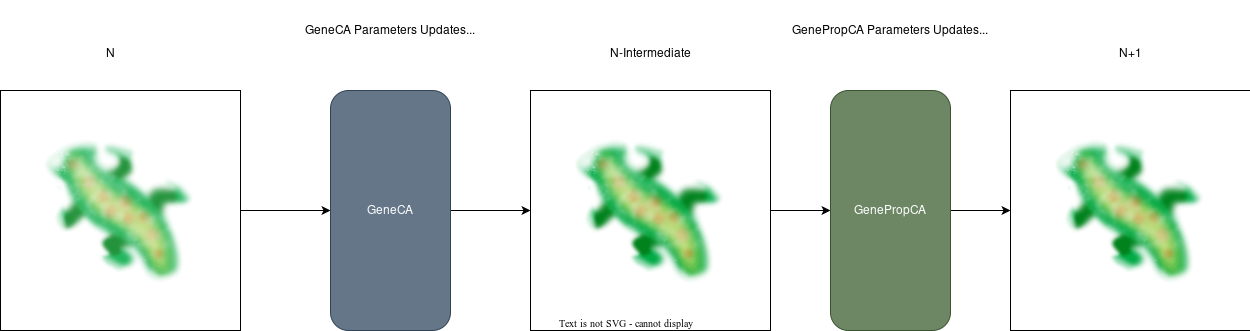}
\caption{One step of the ensemble, GeneCA + GenePropCA}
\label{fig:Ensemble}
\end{figure*}

The EngramNCA ensemble model (Figure \ref{fig:Ensemble}) uses the GeneCA and GenePropCA to produce the final morphology. The GeneCA works without the GenePropCA and will produce the primitives. The GenePropCA is dependent on the GeneCA as its parameters are trained to manipulate the GeneCA dynamics. At step N, the state of the CA is passed to the GeneCA, where it only modifies the visible and hidden channels to produce an intermediate state, the intermediate step is then passed to the GenePropCA, where only the gene encoding channels are modified to produce step N+1. The idea here is for the GenePropCA to learn to exploit the gene-embedded dynamics of the GeneCA to produce more complex morphologies than that found in the GeneCA.
For a formal mathematical definition of the proposed models, including GeneCA, GenePropCA and the ensemble model EngramNCA, together with the training procedure, please refer to the relative section in Appendix 2)

\section{Hierarchical Growth Task}

The traditional NCAs presented in 
\citep{mordvintsev2020growing}
 uses the neural-network parameters to encode the desired behaviour. In effect, the neural-network can be seen as the genome for the morphology, where for each distinct morphology a new NCA needs to be trained. This makes it particularly difficult to have coexisting morphologies, or even just coexisting dynamics, existing on the same lattice, as the multiple NCAs end up interfering with each other.

Our model presents a possible solution to this problem by tying the morphological result to an unchanging seed cell. In this way, we hypothesize that the NCA learns general growing dynamics necessary for any morphology, alongside encoding-specific dynamics for the details of the morphology.

\subsection{Growing Coexisting Primitives}

\begin{figure}[t!]
\centering
\includegraphics[width=0.47\textwidth]{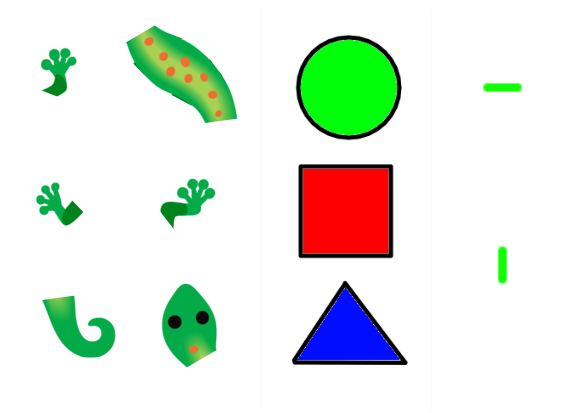}
\caption{Left: All lizard body parts used in training a GeneCA. Middle: All polygons used in training a GeneCA. Right: Vertical and horizontal lines used in training a GeneCA.}
\label{fig:primsAll}
\end{figure}

The first step in the hierarchical growth task was to grow coexisting morphologies. In effect, this means to grow multiple distinct morphologies using the same set of NCA (our GeneCA) parameters. Much like the original growth task as described in 
\citep{mordvintsev2020growing}
, the multiple morphologies should be bounded and stable across time, with their shape unchanging once fully generated. We conducted multiple experiments including; many high-complexity primitives (lizard body parts, Figure \ref{fig:primsAll}, left), fewer medium-complexity primitives (basic polygons, Figure \ref{fig:primsAll}, middle), and fewer low-complexity shapes (vertical and horizontal lines, Figure \ref{fig:primsAll}, right).

\subsection{Growing Morphologies from Primitives}

\begin{figure}[t!]
\centering
\includegraphics[width=0.47\textwidth]{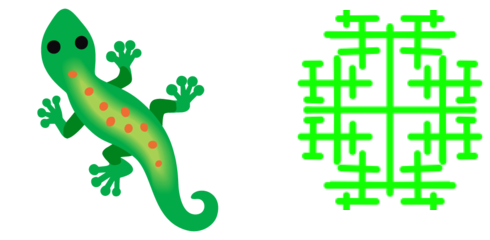}
\caption{Left: Full Lizard morphology used in training the GenePropCA. Right: Fractal-like morphology used in training the GenePropCA .}
\label{fig:lizzFract}
\end{figure}

From a trained GeneCA we then trained the GenePropCA to exploit the latent-space embeddings that the genes now represent. The three different GeneCA mentioned previously where used to test the GenePropCAs capabilities under different circumstances. A high primitive count, low combinatorial problem in growing the lizard (Figure \ref{fig:lizzFract}, left), where there is only one valid combination. A low primitive count, high combinatorial problem in growing a fractal-like shape (Figure \ref{fig:lizzFract}, right) from vertical and horizontal lines. Finally, a combination of the two, by growing a lizard from basic polygons.

\subsection{Growing Coexisting Morphologies}

For final step in the hierarchy, we chose to explore growing multiple morphologies from the same GeneCA and GenePropCA. This was accomplished by extending the model one hierarchy up, and imbuing the GenePropCA with its own meta-genes, that it could see, but neither it nor the GeneCA could modify, and using those meta-genes to encode for entire morphologies.

\section{Training Procedure}

In this section we provide a description of the training procedure, including two distinct training phases, a special pooling method to train multiple primitives at one, and how the GeneCA weights are frozen when training the GenePropCA.

The loss function used for training on all the images is the Pixelwise-MSE, as used in \citep{mordvintsev2020growing}:

\begin{equation}
\resizebox{0.43\textwidth}{!}{$PixelWiseMSE = \frac{1}{H \times W \times C} \sum_{i=0}^{H} \sum_{j=0}^{W} \sum_{k=0  }^{C} \left( I(i, j, k) - \hat{I}(i, j, k) \right)^2$}
\end{equation}

Where $H$, $W$, $W$ are the dimensions of the image, $I$ is the reference image, and $\hat{I}$ is the final state of the NCA.

\subsection{GeneCA Training}

\begin{figure*}[t!]
\centering
\includegraphics[width=0.94\textwidth]{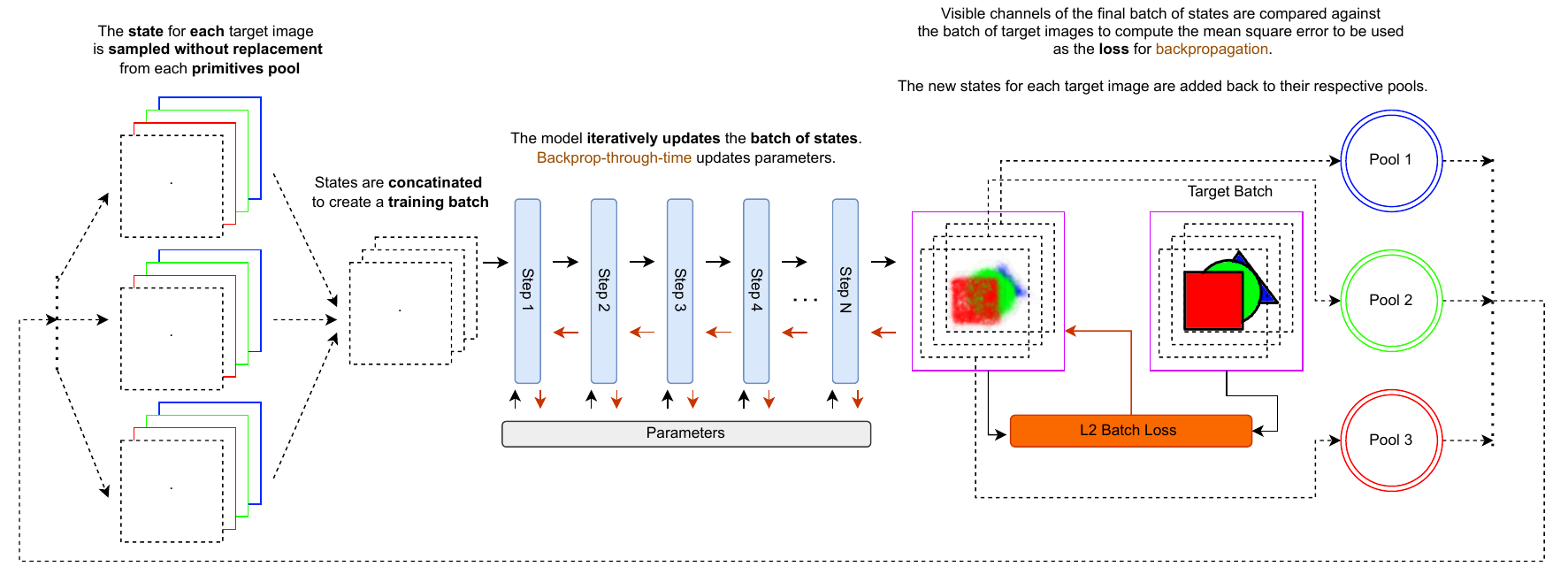}
\caption{Training loop for multi-image batch training}
\label{fig:trainingGene}
\end{figure*}

GeneCA training is similar to the pool-training procedure presented in 
\citep{mordvintsev2020growing}
. However, to enable the gene encoding of multiple primitives, each gene-primitive pair is given its own pool to sample from and update. Additionally, the target image becomes a target batch consisting of n-repetitions of each image, where n is a factor of the batch size. This is done so that multiple images per batch can be considered in the loss. It is important to keep the batch size as a multiple of the number of primitives to avoid training bias towards one primitive. Figure \ref{fig:trainingGene} illustrates the training procedure for batch-multi-image training under the GeneCA scheme.

\subsection{GenePropCA Training}

\begin{figure*}[t!]
\centering
\includegraphics[width=0.94\textwidth]{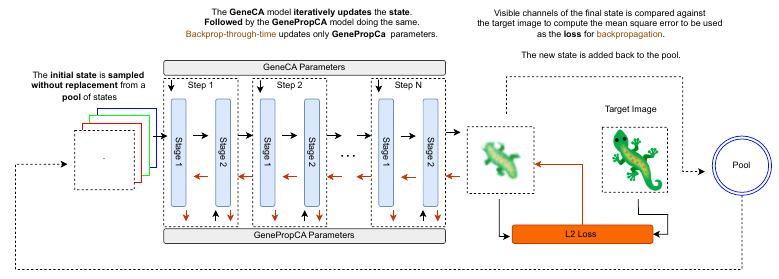}
\caption{Training loop for GenePropCA. The GeneCA weights are frozen, yet still contribute to the gradient computations, making the GenePropCAs behaviour dependent on the GeneCA.}
\label{fig:trainingGeneProp}
\end{figure*}

GenePropCA training uses the same batch training technique as the GeneCA. We typically only train a single morphology, however multiple morphologies can be trained by sacrificing more channels to act as "meta genes". These meta genes are unaffected by both the GeneCA and GenePropCA. During the training of the GenePropCA the GeneCA weights are frozen, though their gradient contribution during the growth process is still used when backpropagating the loss to the GenePropCA. This ensures the GenePropCA's dynamics are conditioned on those of the GeneCA. Figure \ref{fig:trainingGeneProp} shows the GenePropCA training with GeneCA frozen weights.

\section{Results}

The model was tested with different configurations, and below are the results for our experiments. Generally, the NCA has a size of 30 x 30, there are 16 channels in total of which the first four are the alpha value and the three color channels. Of the remaining 12 hidden channels, eight channels were privatized, i.e. could be read but not be altered by the GeneCA.

\subsection{Growing Coexisting Primitives}

Three primitives were chosen as target shapes: A blue square, a green circle and a red triangle. Those primitives have a corresponding one-hot encoding in the private tape. The GeneCA was trained to update the public channels of the CA such that the right primitive is formed around the seed cell. It is important to note that there is no gene propagation at this point, only the update of the public channels that the GeneCA calculates based on the available private and public channels.

During test time, cells of the CA are picked as seed cells, and their private tape is altered.

\begin{figure}[t!]
\centering
\href{https://etimush.github.io/EngramNCA/#growing-prims}{
    \begin{subfigure}[b]{0.15\textwidth}
        \centering
         \includegraphics[width=\textwidth]{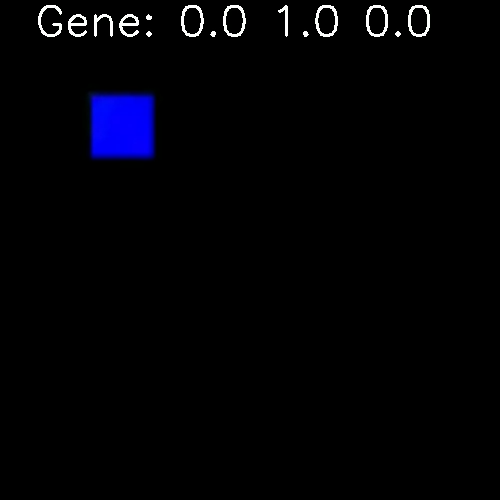}
         \caption{step 50}
         \label{step 50}
    \end{subfigure}
    \begin{subfigure}[b]{0.15\textwidth}
        \centering
         \includegraphics[width=\textwidth]{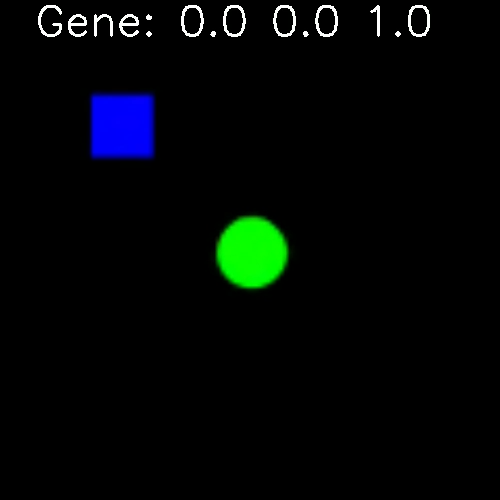}
         \caption{step 100}
         \label{fig:y equals x}
    \end{subfigure}
    \begin{subfigure}[b]{0.15\textwidth}
        \centering
         \includegraphics[width=\textwidth]{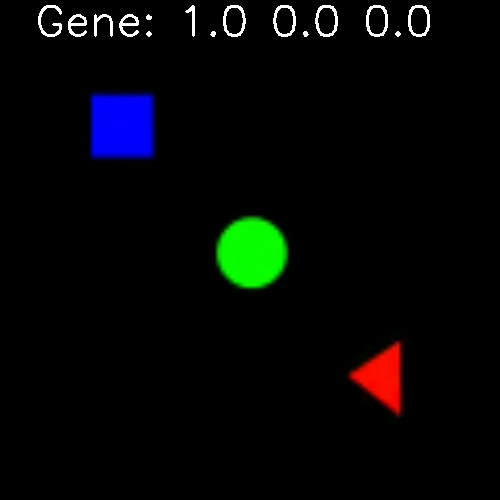}
         \caption{step 150}
         \label{fig:y equals x}
    \end{subfigure}
    }
\caption{Growing simple primitives.}
\label{fig:growingPrims}
\end{figure}

In the top of Figure \ref{fig:growingPrims} we show how the binary encodings of the three primitives are assigned to three different cells in the CA, with the distance between them being great enough to avoid interactions between cells belonging to different primitives. Starting from the seed cell, the shapes and their color form over a number of update steps. Afterwards, they remain stable.

\begin{figure}[t!]
\centering
\href{https://etimush.github.io/EngramNCA/#individuality-demo}{
    \includegraphics[width=0.3\textwidth]{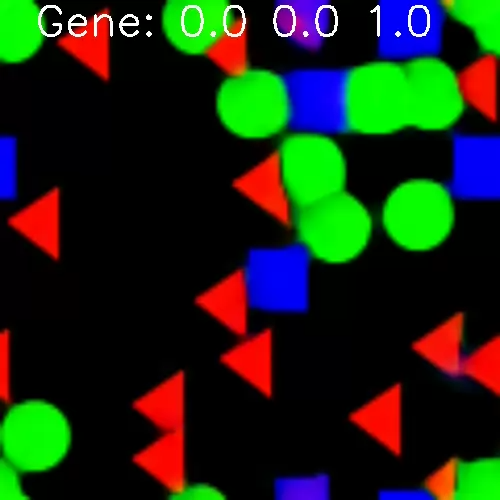}
}
\caption{Untrained behaviour: primitives displaying strong individuality.}
\label{fig:individuals}
\end{figure}

Figure \ref{fig:individuals} shows what happens if seed cells are placed in such proximity to each other that the resulting primitives would normally overlap. As can be seen, the shapes remain largely distinct. Especially shapes of the same kind have neither overlap nor mixing, and instead, a border can be visually seen. Circles placed in proximity to each other can, for example, be seen to form a comb-like structure (top of Figure \ref{fig:individuals}). If shapes of different kinds meet, the distinction is not always as clear and some interactions between different primitives can lead to a mixing of colors, as can be seen in the bottom of Figure \ref{fig:individuals}. Still, a distinction can be made, and no primitive takes over another primitive. This happens without such cases occurring during training.

\begin{figure}[t!]
\centering
\href{https://etimush.github.io/EngramNCA/#primitive-mixing}{
\includegraphics[width=0.3\textwidth]{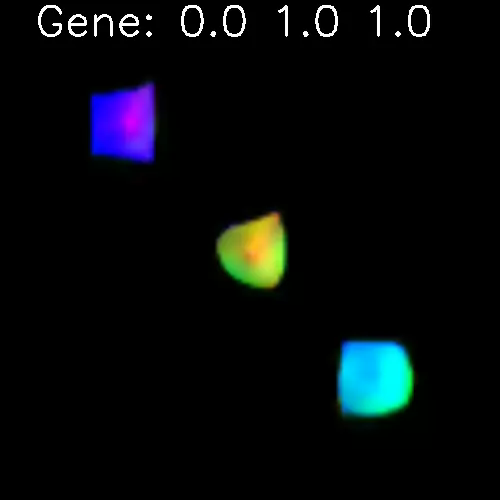}
}
\caption{Untrained behaviour: out-of-training generative capabilities from mixing genes.}
\label{fig:mixedprims}
\end{figure}

Only the three one-hot encodings were used during training, but it is possible to deviate from this during test time and to use different binary arrays as the private encodings of the seed cell. This essentially means that there is a mixture of genes. As can be seen in Figure \ref{fig:mixedprims}, this leads to the growth of corresponding mixed forms. It can be seen that the shape of those forms is a distorted overlap of the primitives, while the color ranges from the pure color of the primitives to the mix of the two colors. The primitives dominate the developing form in those areas where the other primitive is not growing, hence the combination of circle and square leads to most mixing. After a developmental period, the forms remain stable. This stability remains for any form of encoding mixing and seems to be bound by the maximum size of the primitives.

\begin{figure}[t!]
\centering
\href{https://etimush.github.io/EngramNCA/#body-part-prims}{
\includegraphics[width=0.3\textwidth]{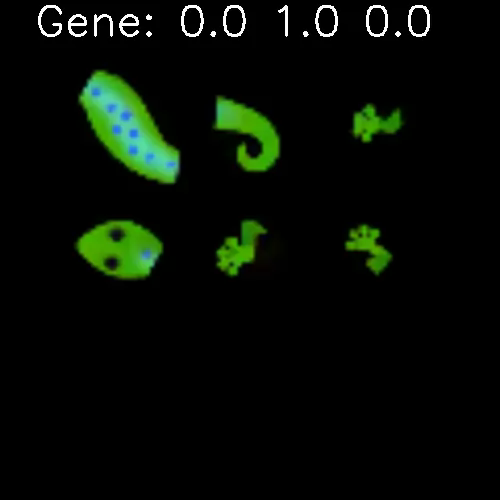}
}
\caption{Lizard body parts primitives.}
\label{fig:bodypartPrimsgrowing}
\end{figure}

The experiments were repeated with body parts of the lizard image as primitives as can be seen in Figure \ref{fig:bodypartPrimsgrowing}. As there were more than three parts, the encoding is no longer one-hot. As before, placing a seed cell with the corresponding encoding leads to the growth and stable persistence of primitives, even though those have more complex shapes and colorings.

\begin{figure}[t!]
\centering
\href{https://etimush.github.io/EngramNCA/#line-prims}{
\includegraphics[width=0.3\textwidth]{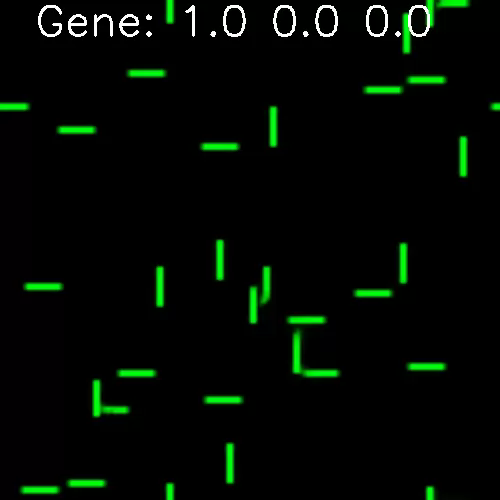}
}
\caption{Horizontal and vertical line primitives.}
\label{fig:lineprims}
\end{figure}

Lastly, very simple unicolor vertical and horizontal lines were used as primitive, as is depicted in Figure \ref{fig:lineprims}. This is the setup for later experiments.

\subsection{Growing Morphologies from Primitives}

In the next experimental step, the GenePropCA was included. The attempt was to grow whole morphologies while the GeneCA weights were frozen on growing primitives.

\begin{figure}[t!]
\centering
\href{https://etimush.github.io/EngramNCA/#liz-from-parts}{
    \begin{subfigure}[b]{0.15\textwidth}
        \centering
         \includegraphics[width=\textwidth]{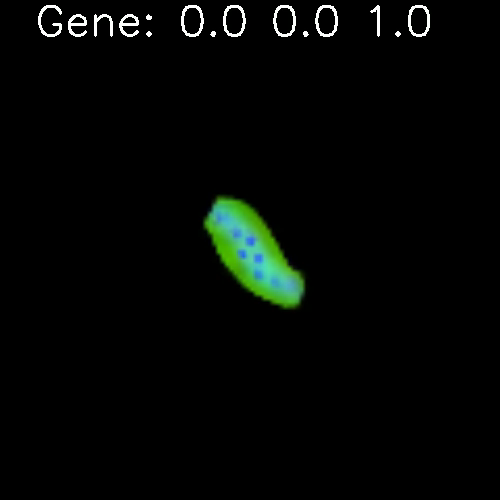}
         \caption{step 16}
         \label{fig:y equals x}
    \end{subfigure}
    \begin{subfigure}[b]{0.15\textwidth}
        \centering
         \includegraphics[width=\textwidth]{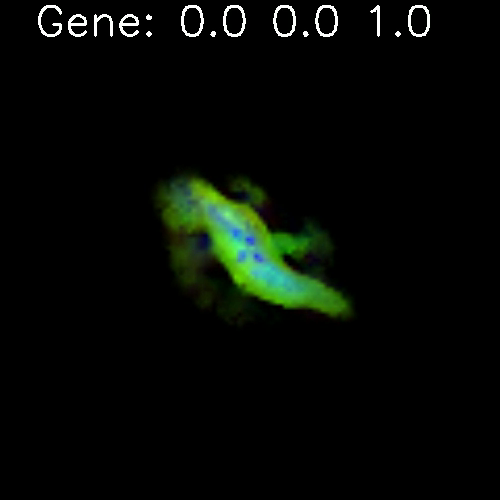}
         \caption{step 32}
         \label{fig:y equals x}
    \end{subfigure}
    \begin{subfigure}[b]{0.15\textwidth}
        \centering
         \includegraphics[width=\textwidth]{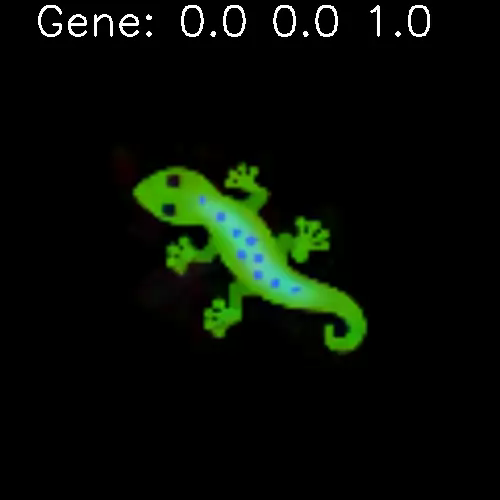}
         \caption{step 64}
         \label{fig:y equals x}
    \end{subfigure}
    }
\caption{Growing the lizard morphology from the torso.}
\label{fig:growingfromtorse}
\end{figure}

In the lizard example, the GenePropCA was trained as described above, with the initial seed cell containing the gene encoding for the torso primitive and the GeneCA being trained on primitive body parts of the whole lizard (legs, head, torso, tail). As can be seen in Figure \ref{fig:growingfromtorse}, the torso of the lizard is grown first, but the CA then continues to grow the limbs, tail, and head. Deactivating the GenePropCA after the lizard is finished growing does not lead to a collapse, rather the lizard maintains its shape under the control of the GeneCA.

\begin{figure}[t!]
\centering
\href{https://etimush.github.io/EngramNCA/#liz-from-shapes}{
\begin{subfigure}[b]{0.15\textwidth}
        \centering
         \includegraphics[width=\textwidth]{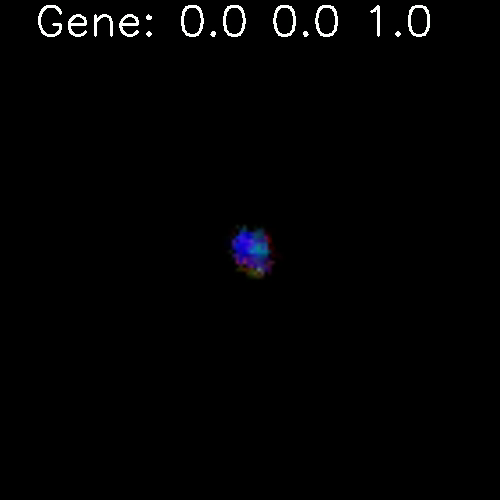}
         \caption{step 10}
         \label{fig:y equals x}
    \end{subfigure}
    \begin{subfigure}[b]{0.15\textwidth}
        \centering
         \includegraphics[width=\textwidth]{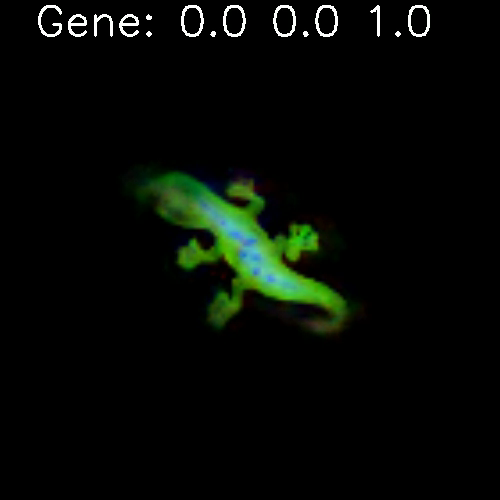}
         \caption{step 40}
         \label{fig:y equals x}
    \end{subfigure}
    \begin{subfigure}[b]{0.15\textwidth}
        \centering
         \includegraphics[width=\textwidth]{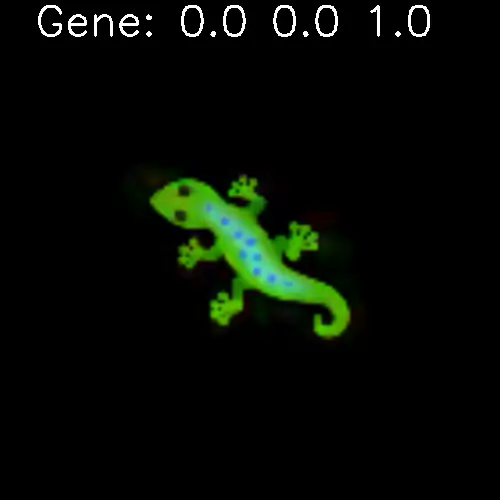}
         \caption{step 64}
         \label{fig:y equals x}
    \end{subfigure}
    }
\caption{Growing the lizard from basic polygon morphologies.}
\label{fig:growingfrompoly}
\end{figure}

\begin{figure}[t!]
\centering
\href{https://etimush.github.io/EngramNCA/#liz-diff-start}{
    \begin{subfigure}[b]{0.15\textwidth}
        \centering
         \includegraphics[width=\textwidth]{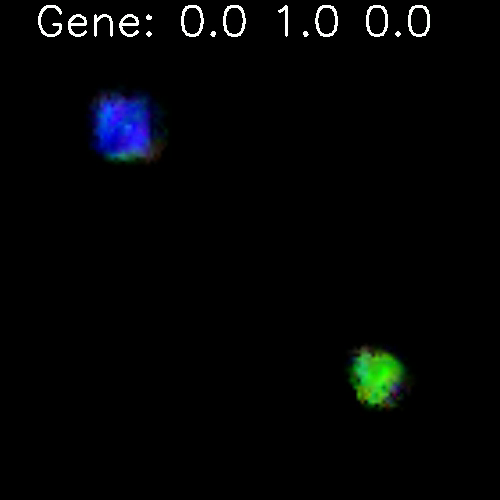}
         \caption{step 16}
         \label{fig:y equals x}
    \end{subfigure}
    \begin{subfigure}[b]{0.15\textwidth}
        \centering
         \includegraphics[width=\textwidth]{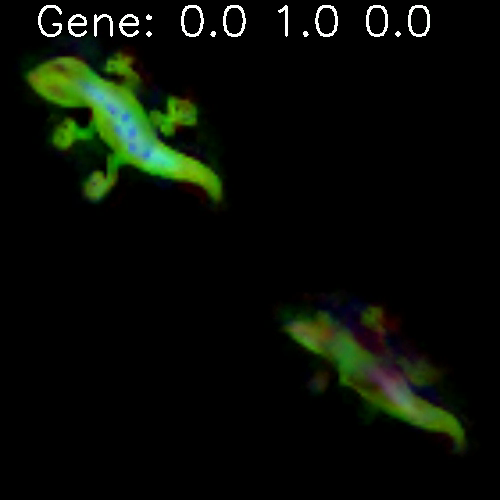}
         \caption{step 32}
         \label{fig:y equals x}
    \end{subfigure}
    \begin{subfigure}[b]{0.15\textwidth}
        \centering
         \includegraphics[width=\textwidth]{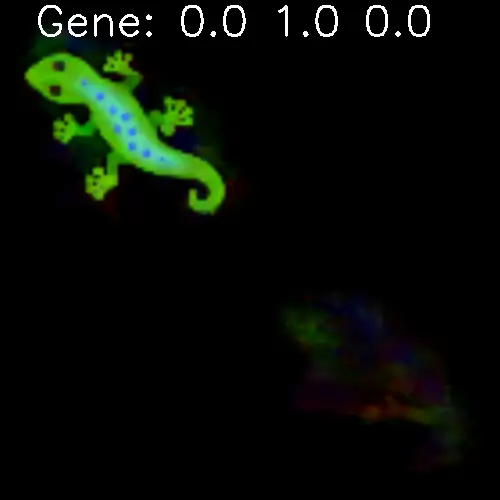}
         \caption{step 64}
         \label{fig:y equals x}
    \end{subfigure}
    }
\caption{Growing the lizard from basic polygon morphologies. The lizard will not grow from a primitive it was not trained to grow from.}
\label{fig:growing2poly}
\end{figure}

The experiment was repeated, this time choosing the circle, square and triangle as primitives and the blue circle as the seed cell during the GenePropCA training. This was successful, as can be seen in Figure \ref{fig:growingfrompoly}. Even if the GeneCA is not trained on all primitives occurring in the final lizard or even on primitives not derived from the target image at all, the GenePropCA is able to generalize and facilitate gene propagation that leads to the growth of the lizard, albeit with less quality as in the case described above for Figure \ref{fig:growingfromtorse}. Figure \ref{fig:growing2poly} shows that the choice of the seed cell does indeed matter, as the lizard will only grow successfully if the exact same seed as in the training of the GenePropCA is used.

Further experiments showed that different seeds, which were not used as primitives during the training of the GeneCA, can be used to train the GenePropCA to grow a lizard.

\begin{figure}[t!]
\centering
\href{https://etimush.github.io/EngramNCA/#grow-frac}{
    \begin{subfigure}[b]{0.15\textwidth}
        \centering
         \includegraphics[width=\textwidth]{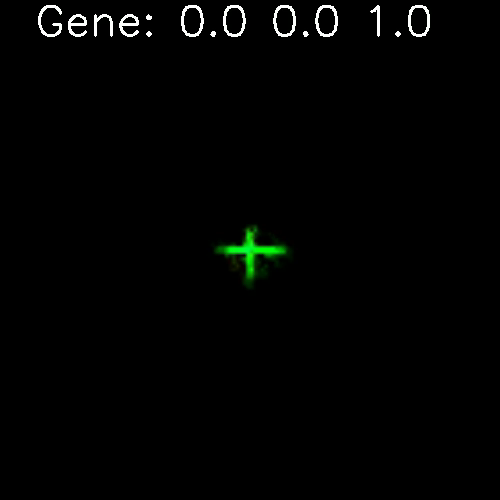}
         \caption{step 16}
         \label{fig:y equals x}
    \end{subfigure}
    \begin{subfigure}[b]{0.15\textwidth}
        \centering
         \includegraphics[width=\textwidth]{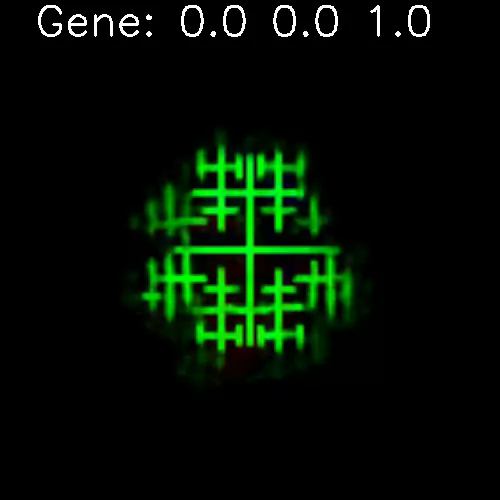}
         \caption{step 32}
         \label{fig:y equals x}
    \end{subfigure}
    \begin{subfigure}[b]{0.15\textwidth}
        \centering
         \includegraphics[width=\textwidth]{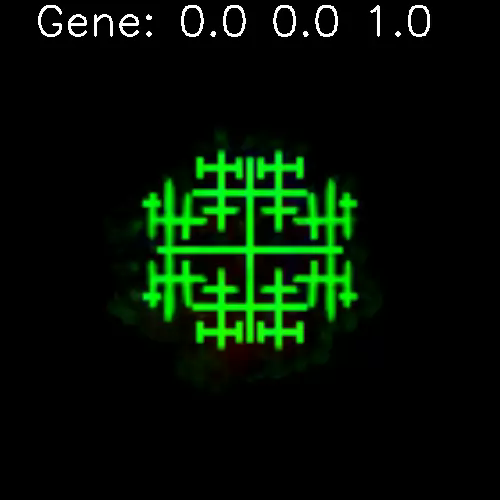}
         \caption{step 64}
         \label{fig:y equals x}
    \end{subfigure}
    }
\caption{Growing a fractal-like shape from simple vertical and horizontal primitives.}
\label{fig:fractallike}
\end{figure}

Figure \ref{fig:fractallike} depicts an experiment in which the simple primitives of unicolor vertical and horizontal lines are combined into a fractal-like shape with satisfying performance. This shows that the model can combine simple and few primitives into more complex forms consisting of many primitives in addition to combining complex and many primitives into less complex (complexity in the sense of the combination of primitives) forms consisting of few primitives, as was done before with the lizard.

\subsection{Growing Coexisting Morphologies from Primitives}

The final experiment in growing morphologies was to test whether the GenePropCA could additionally grow multiple morphologies. This was done by adding an extra channel as a meta-gene, that neither the GeneCA nor GenePropCA could modify and training the GenePropCA to grow separate morphologies depending on the meta-gene encoding and the starting seed cell encoding.

\begin{figure}[t!]
\centering
\href{https://etimush.github.io/EngramNCA/#2-morph}{
    \begin{subfigure}[b]{0.15\textwidth}
        \centering
         \includegraphics[width=\textwidth]{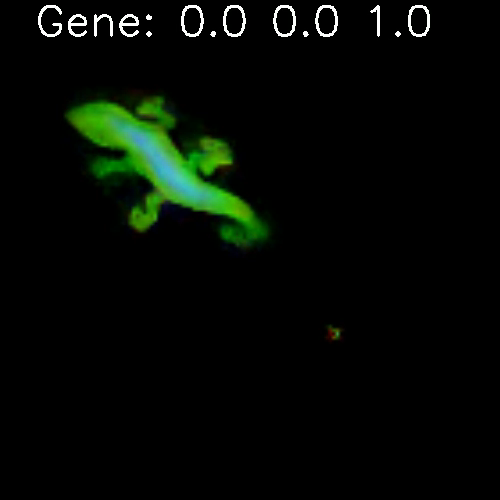}
         \caption{step 100}
         \label{fig:y equals x}
    \end{subfigure}
    \begin{subfigure}[b]{0.15\textwidth}
        \centering
         \includegraphics[width=\textwidth]{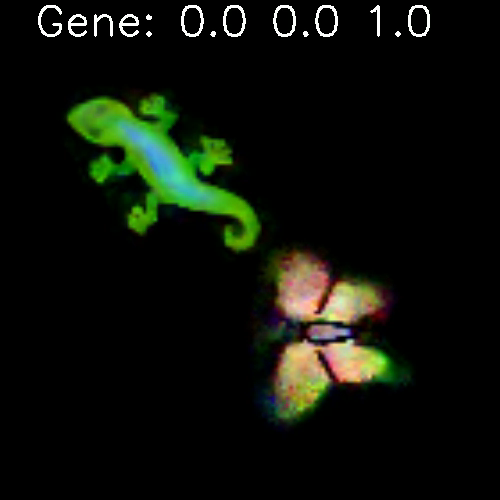}
         \caption{step 150}
         \label{fig:y equals x}
    \end{subfigure}
    \begin{subfigure}[b]{0.15\textwidth}
        \centering
         \includegraphics[width=\textwidth]{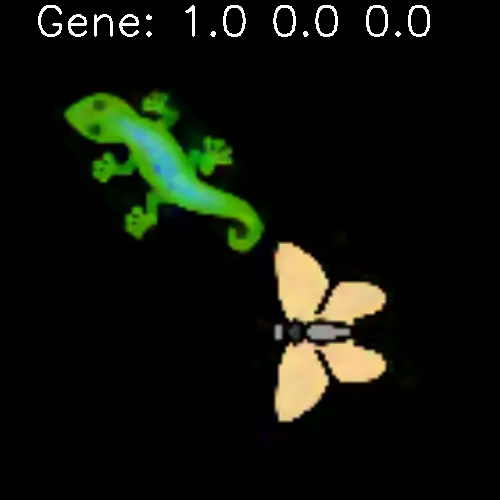}
         \caption{step 200}
         \label{fig:y equals x}
    \end{subfigure}
    }
\caption{Growing a lizard and a butterfly with the same GeneCA and GenePropCA. The quality of reproduction is reduced.}
\label{fig:2morpho}
\end{figure}

Figure \ref{fig:2morpho} shows the GenePropCA growing both a lizard morphology and a butterfly by exploiting the same GeneCA gene encodings. The overall quality of reproduction of the lizard has diminished, whilst the butterfly, a much simpler morphology, was reproduced better.

\subsection{Moving Primitives}

\begin{figure}[t!]
\centering
\href{https://etimush.github.io/EngramNCA/#originalLenia}{
    \begin{subfigure}[b]{0.15\textwidth}
        \centering
         \includegraphics[width=\textwidth]{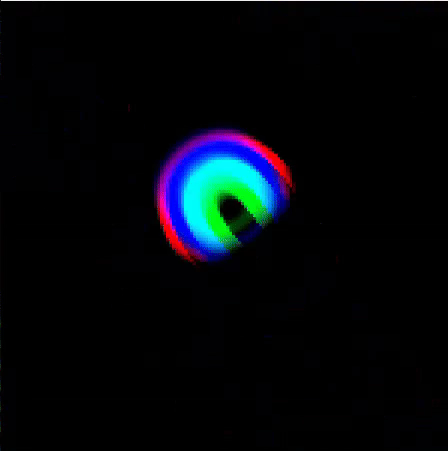}
         \caption{step 10}
         \label{fig:y equals x}
    \end{subfigure}
    \begin{subfigure}[b]{0.15\textwidth}
        \centering
         \includegraphics[width=\textwidth]{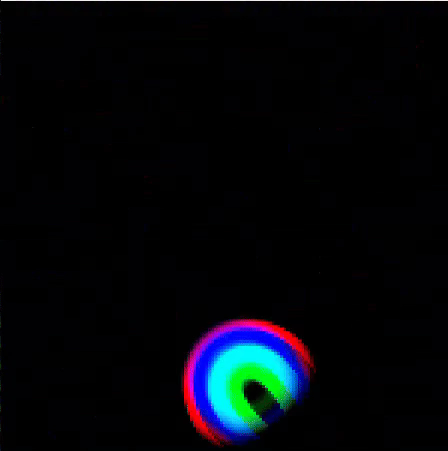}
         \caption{step 100}
         \label{fig:y equals x}
    \end{subfigure}
    \begin{subfigure}[b]{0.15\textwidth}
        \centering
         \includegraphics[width=\textwidth]{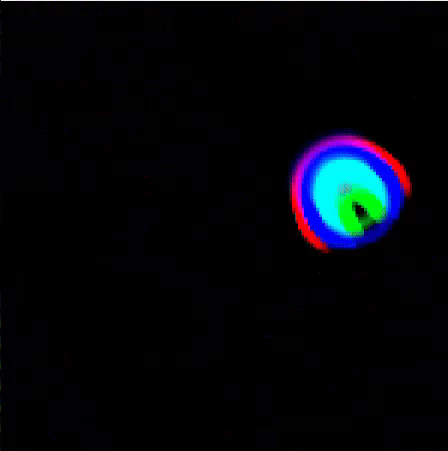}
         \caption{step 400}
         \label{fig:y equals x}
    \end{subfigure}
    }
\caption{Original Lenia video.}
\label{fig:lenia}
\end{figure}

In contrast to the relatively simple task of growing morphologies, we wanted to test if the GenePropCA could exploit the embedding and learn the dynamics of a complex moving system. For this we chose to replicate the dynamics of a Lenia 
\citep{Lenia2019}
 glider. Lenia itself is a form of continuous CA that can exhibit remarkable complexity. We chose a glider (Figure \ref{fig:lenia}) whose dynamics were non-trivial, where learning to replicate it would take more than simply memorizing a simple translation in space. The GeneCA was trained on the first frame of the Lenia video. The GenePropCA was trained on a quasi-curriculum-style learning, where at certain checkpoints, additional frames of the video are inserted into the training set until the five-hundreds frame. It is always trained from the first frame to the last frame of the training set in order, and loss was accumulated across all frames before backpropagation. The frame-by-frame loss accumulation is important here since -unlike growing morphologies- the time evolution of the NCA from state to state is important. As a comparison, a standard NCA was trained using the same method, and the parameter count of the standard NCA was selected such that it corresponds to the sum of total parameters between the GeneCA and the GenePropCA.

\begin{figure}[t!]
\centering
\href{https://etimush.github.io/EngramNCA/#standardNCA}{
    \begin{subfigure}[b]{0.15\textwidth}
        \centering
         \includegraphics[width=\textwidth]{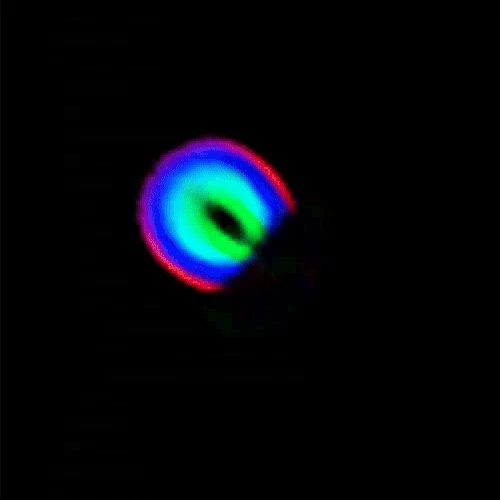}
         \caption{step 10}
         \label{fig:y equals x}
    \end{subfigure}
    \begin{subfigure}[b]{0.15\textwidth}
        \centering
         \includegraphics[width=\textwidth]{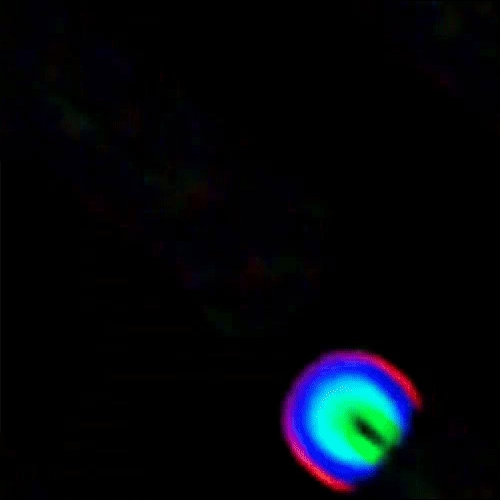}
         \caption{step 100}
         \label{fig:y equals x}
    \end{subfigure}
    \begin{subfigure}[b]{0.15\textwidth}
        \centering
         \includegraphics[width=\textwidth]{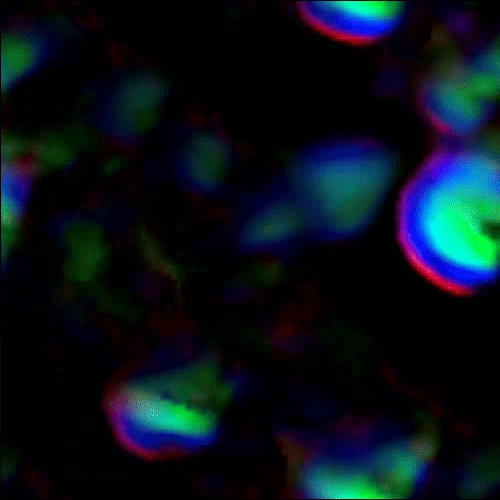}
         \caption{step 400}
         \label{fig:y equals x}
    \end{subfigure}
    }
\caption{Standard NCA trained to replicate the video.}
\label{fig:standardca}
\end{figure}

Figure \ref{fig:standardca} shows the result of training a standard NCA on five-hundred frames of the Lenia video. As can be seen, the standard NCA somewhat replicates the dynamics until a certain point, where the simulation falls apart. Possibly due to the NCA only learning to replicate the exact frames it saw and not learning the dynamics.

\begin{figure}[t!]
\centering
\href{https://etimush.github.io/EngramNCA/#movingGene}{
    \begin{subfigure}[b]{0.15\textwidth}
        \centering
         \includegraphics[width=\textwidth]{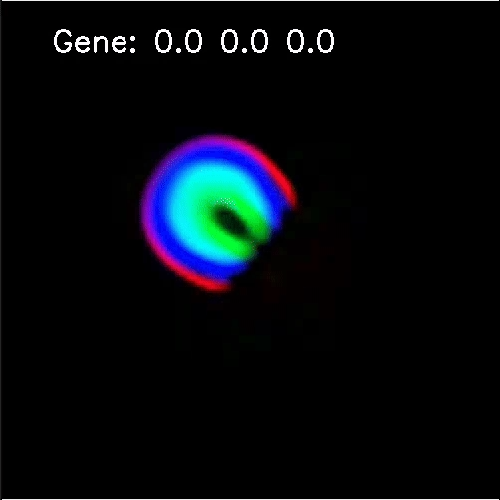}
         \caption{step 10}
         \label{fig:y equals x}
    \end{subfigure}
    \begin{subfigure}[b]{0.15\textwidth}
        \centering
         \includegraphics[width=\textwidth]{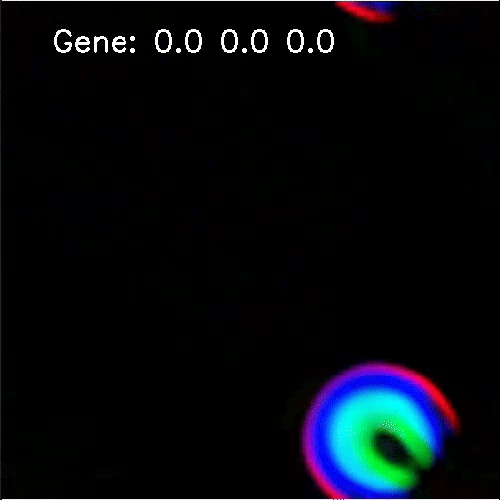}
         \caption{step 100}
         \label{fig:y equals x}
    \end{subfigure}
    \begin{subfigure}[b]{0.15\textwidth}
        \centering
         \includegraphics[width=\textwidth]{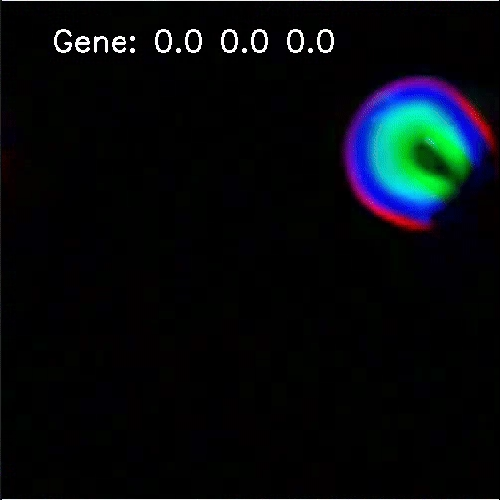}
         \caption{step 400}
         \label{fig:y equals x}
    \end{subfigure}
    }
\caption{Our method trained to replicate the video.}
\label{fig:movingGene}
\end{figure}

(Figure \ref{fig:movingGene}) shows the result of training the GeneCA and GenePropCA on five-hundred frames of the Lenia video. As can be seen, our method yields visually closer results to the Lenia video. Additionally, the morphology is stable past the five-hundred frame mark, exhibiting signs of having learned more general dynamics that are stable in time. Though not a perfect replication of the Lenia glider, the results are decent.

\section{Summary and Discussion}

The prevailing paradigm in neuroscience has long held that memory is predominantly stored and transmitted through synaptic plasticity—the so‐called “synaptic dogma” \citep{Kandel2001}. This view posits that modifications in synaptic strength are the primary means by which information is encoded in neural circuits. However, a growing body of experimental evidence suggests that memory storage may also involve intracellular processes and molecular mechanisms. For instance, studies in aplysia have demonstrated that injection of RNA extracted from trained animals into naïve counterparts can induce memory‐like changes, implying that molecular substrates may contribute to memory formation beyond traditional synaptic modifications \citep{bedecarrats2018}. Similarly, research on planaria has shown that decapitated individuals can retain learned behaviors after head regeneration, further suggesting that memory may reside in non‐synaptic, cellular compartments \citep{McConnell1962, Shomrat2013}. These findings have significant implications for both biological and computational models of memory. Whereas classical models emphasize network-level interactions and synaptic weight adjustments, recent work has increasingly focused on the roles of gene expression and intracellular signaling cascades in long-term memory storage \citep{Kandel2014}. Insights from developmental biology also support this broader perspective; gene regulatory networks, for example, orchestrate complex morphogenetic processes \citep{Davidson2006}, underscoring the potential for multi-timescale information transfer that integrates fast synaptic signaling with slower, gene-mediated processes.

Inspired by these biological insights, emerging computational frameworks—such as the one proposed here and named EngramNCA—are exploring dual modes of information storage. In these models, the system is endowed with a bifurcated state representation: a publicly accessible channel that mirrors synaptic activity and a private channel that represents intracellular memory. This dual-channel approach reflects the biological separation between immediate synaptic signaling and slower molecular memory transfer, offering a more nuanced basis for capturing complex developmental and memory dynamics. Integrating biological principles of memory into computational models may ultimately pave the way for more robust and adaptable systems. 

Our results suggest that having fixed memory portions in the cells makes for more stable expression of learned patterns. Moreover, our results show that the learned dynamic behavior of Lenia could be expressed more accurately using a private memory. This observation supports the proposed mechanism as a possible explanation for how memory could be stored in real bodies. More importantly, it also shows that cell memory could be a promising approach to having more stable distributed systems in AI. Stability is an open problem in AI, which is typically mitigated by pool-training. One avenue for future work is to use EngramNCA cell memory mechanisms when pool-training is not applicable. Additionally, we suggest that EngramNCA could suit more complex datasets such as the Abstraction and Reasoning Corpus (ARC) \citep{chollet2019measure}. 

Further, the EngramNCA model could provide valuable insights for neuroscience by offering a computational framework that aligns with emerging biological evidence of memory mechanisms beyond the traditional synaptic dogma. By demonstrating how systems with both publicly visible states and private internal memory can develop more stable patterns, hierarchical organization, and robust information transfer, this model suggests potential mechanisms for biological phenomena like memory retention after tissue regeneration or RNA-mediated memory transfer. Specifically, the model illustrates how intracellular memory components might work alongside synaptic connections to create more resilient and multifaceted memory systems—potentially explaining why planaria can maintain memories after decapitation or why RNA transfer can induce learned behaviors. This computational approach could guide future experimental design in neuroscience by predicting how cellular-level memory mechanisms might interact across different timescales and spatial organizations, ultimately leading to a more nuanced understanding of memory encoding, storage, and retrieval in biological systems.

\section{Acknowledgments}
This work was partially supported by The Digital Society strategic initiative at Øsftold University College.

\footnotesize
\bibliographystyle{apalike}
\bibliography{main} 

\section{Appendix 1}

This section details the preliminary experiments conducted to determine the deterioration in NCA image reconstruction quality as channels are privatized. We measured the loss curves for reconstruction of a lizard emoji, alongside the regrowth capabilities of the NCA to damage. For all experiments the training was repeated once on a set seed for each privatization level, the regrowth experiments where repeated one-hundred times with the damage locations being fixed and identical for each privatization level.
\\
\\
For these experiments, we tested three different NCA types:
\begin{itemize}
  \item \textbf{DummyVCA}: Where the sensing kernels channel dimension match the unmasked NCA channels. However, any kernel involved in sensing neighbourhoods (such as sobel filters) would receive a dummy vector of all zeros for in lieu of the privatized channels.

  \item \textbf{MaskedCA}: Where the sensing kernels channel dimension match the unmasked NCA channels. However, the channels of the output tensor corresponding to the privatized channels where masked with zeros after the sensing convolution.
  
  \item \textbf{ReducedCA}: Where the sensing kernels channel dimension where reduced to not include the privatized channels, and a truncated state excluding the privatized channels is passed to the sensing kernels. The state was later re-expanded with zeros t mathc the unmasked NCA channel dimensions.
\end{itemize}

The loss function used for training and measuring the reconstruction was the PixelwiseMSE:

\begin{equation}
\resizebox{0.43\textwidth}{!}{$PixelWiseMSE = \frac{1}{H \times W \times C} \sum_{i=0}^{H} \sum_{j=0}^{W} \sum_{k=0  }^{C} \left( I(i, j, k) - \hat{I}(i, j, k) \right)^2$}
\end{equation}

Where $H$, $W$, $W$ are the dimensions of the image, $I$ is the reference image, and $\hat{I}$ is the final state of the NCA.

\subsection{Preliminary experiment results}

\begin{figure}[t!]
\centering
\includegraphics[width=0.47\textwidth]{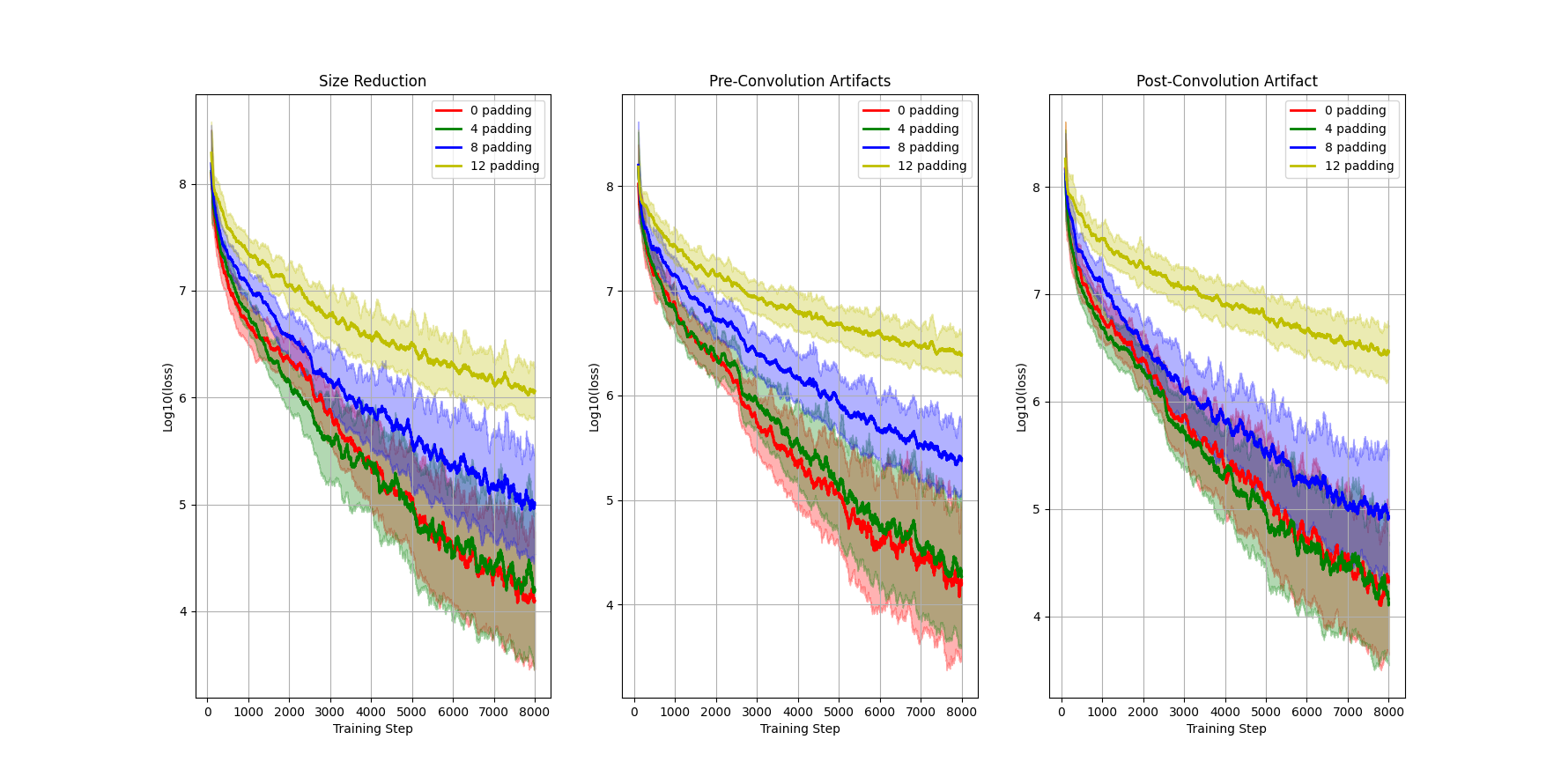}
\caption{Comparison of the loss curves of the three models (left to right) with different levels of privatization (0,4,8,12, top to bottom. For privatization levels 0 and 4 the three models are comparable, while at privatization level 8 the DummyVCA performed considerably worse.Finally, the ReducedCA performed best at privatization level 12.}
\label{fig:inter_model}
\end{figure}

Figure \ref{fig:inter_model} shows the loss curves of the three models (left to right) with different levels of privatization (0,4,8,12, top to bottom). For privatization levels 0 and 4 the three models are comparable, while at privatization level 8 the DummyVCA performed considerably worse.Finally, the ReducedCA performed best at privatization level 12. From privatization level 0 to 4 there seems to be no discernible change in the loss curves, afterward however, as more channels get privatized the models struggle to learn.\\

\begin{figure}[t!]
\centering
\includegraphics[width=0.47\textwidth]{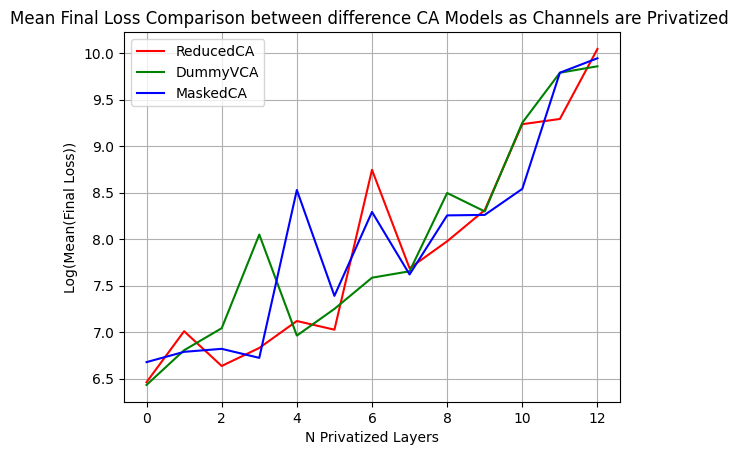}
\caption{Mean final loss as compared to number of channels privatized for each of the three models. As channels are privatized, it can be seen that the reconstruction quality of the original image (mean final loss) decreases.
}
\label{fig:inter_model_loss_fine}
\end{figure}

Figure \ref{fig:inter_model_loss_fine} shows how the reconstruction quality (the mean final loss, on a running average of 10) on the lizard emoji decreases as channels are privatized. The data is noisy, thus no proper comparison can be done between the three models, they can be considered to act the same. \\ 

\begin{figure}[t!]
\centering
\includegraphics[width=0.47\textwidth]{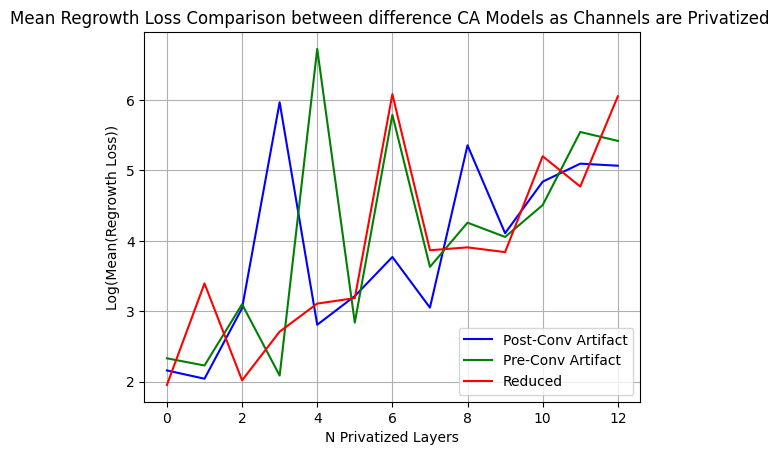}
\caption{Mean regrowth loss as compared to number of channels privatized for each of the three models. As channels are privatized, it can be seen that the NCA struggles to regrow back to the original image.
}
\label{fig:regrowth_loss}
\end{figure}

Figure \ref{fig:regrowth_loss} shows the mean regrowth loss of the three models as they attempt to reconstruct the damaged lizard. As with the mean final loss, regrowth quality decreases as the number of channels are privatized. The data is once again noisy, even more so here. This is most likely due to the asynchronous nature of NCAs. \\

From the results of the preliminary experiments, we concluded that there is indeed deterioration in the quality of image reconstruction from the NCA as channels are privatized. However, the choice of architecture does not play a big role in the level of performance loss. Thus, we chose to go with the MaskedCA version for channel reduction as it was the easiest to implement.

\section{Appendix 2}
\subsection{Mathematical details of the proposed models}
In this section we define the proposed models with mathematical detail.
\subsubsection{Cellular automaton framework}

We formalize our EngramNCA as a grid of cells $C \in \mathbb{R}^{H \times W \times N}$, where $H$ and $W$ are the grid dimensions, and $N$ is the total number of channels. Each cell $c_{i,j}$ at position $(i,j)$ contains a state vector that is partitioned into three components:

$$c_{i,j} = [v_{i,j}, h_{i,j}, g_{i,j}]$$

where $v_{i,j} \in \mathbb{R}^4$ represents the visible RGB-$\alpha$ channels, $h_{i,j} \in \mathbb{R}^{n_h}$ represents the public hidden channels, and $g_{i,j} \in \mathbb{R}^{n_g}$ represents the private gene channels. The dimensions satisfy $4 + n_h + n_g = N$.

\subsubsection{GeneCA architecture}

GeneCA updates the public channels (visible and hidden) while preserving the private gene channels. For each cell $c_{i,j}$, the update rule is defined as:

$$[v_{i,j}^{t+1}, h_{i,j}^{t+1}] = [v_{i,j}^t, h_{i,j}^t] + \phi_\text{GeneCA}(\mathcal{P}(c_{i,j}^t), g_{i,j}^t)$$
$$g_{i,j}^{t+1} = g_{i,j}^t$$

where $\mathcal{P}(c_{i,j}^t)$ represents the perception vector derived from the cell's neighborhood, and $\phi_\text{GeneCA}$ is a neural network that computes the update. The perception function $\mathcal{P}$ applies convolution kernels to the grid:

$$\mathcal{P}(c_{i,j}^t) = [\text{Identity}(c_{i,j}^t), \text{Sobel}_x(c_{i,j}^t), \text{Sobel}_y(c_{i,j}^t), \text{Laplacian}(c_{i,j}^t)]$$

where Identity, Sobel$_x$, Sobel$_y$, and Laplacian are convolution filters applied only to the visible and hidden channels. The neural network $\phi_\text{GeneCA}$ is structured as:

$$\phi_\text{GeneCA}(\mathcal{P}, g) = (W_2 \cdot \text{ReLU}(W_1 \cdot [\mathcal{P}, g] + b_1)) \cdot u_m \cdot l_m$$

where $W_1, W_2, b_1$ are learnable parameters and $u_m, l_m$ are the asynchronous update mask and cell living mask respectively. 

\subsubsection{GenePropCA architecture}

The GenePropCA updates only the gene channels while preserving the public channels:

$$[v_{i,j}^{t+1}, h_{i,j}^{t+1}] = [v_{i,j}^t, h_{i,j}^t]$$
$$g_{i,j}^{t+1} = g_{i,j}^t + \psi_\text{GenePropCA}(\mathcal{P}(c_{i,j}^t), g_{i,j}^t)$$

where $\psi_\text{GenePropCA}$ is a neural network with a similar architecture to $\phi_\text{GeneCA}$ but outputs updates for gene channels only:

$$\psi_\text{GenePropCA}(\mathcal{P}, g) = (V_2 \cdot \text{ReLU}(V_1 \cdot [\mathcal{P}, g] + d_1)) \cdot u_m \cdot l_m$$

where $V_1, V_2, d_1$ are learnable parameters and $u_m, l_m$ are the asynchronous update mask and cell living mask respectively.

\subsubsection{EngramNCA ensemble}

The EngramNCA ensemble combines both models in sequence. For a single update step:

$$c_{i,j}^{t+\frac{1}{2}} = [v_{i,j}^t + \Delta v_{i,j}^t, h_{i,j}^t + \Delta h_{i,j}^t, g_{i,j}^t]$$

where $[\Delta v_{i,j}^t, \Delta h_{i,j}^t] = \phi_\text{GeneCA}(\mathcal{P}(c_{i,j}^t), g_{i,j}^t)$, followed by:

$$c_{i,j}^{t+1} = [v_{i,j}^{t+\frac{1}{2}}, h_{i,j}^{t+\frac{1}{2}}, g_{i,j}^t + \Delta g_{i,j}^t]$$

where $\Delta g_{i,j}^t = \psi_\text{GenePropCA}(\mathcal{P}(c_{i,j}^{t+\frac{1}{2}}), g_{i,j}^t)$.

\subsubsection{Training procedure}

The GeneCA is trained first with frozen gene channels. For each training iteration, we:

1. Sample a batch of $B$ cells from pools corresponding to $K$ different primitive morphologies
2. Initialize the gene channels of each seed cell with a unique binary encoding $E_k \in \{0,1\}^{n_g}$ for primitive $k$
3. Run the GeneCA for $T$ steps to grow the morphologies
4. Compute the loss using pixelwise MSE between the final visible channels and target images:

$$\mathcal{L}_\text{GeneCA} = \frac{1}{BHW} \sum_{b=1}^B \sum_{i=1}^H \sum_{j=1}^W ||v_{i,j}^T - \hat{v}_{i,j}||_2^2$$

After training GeneCA, its weights are frozen, and GenePropCA is trained to propagate and modify gene information to grow complex morphologies. The same loss function is used, but with target images representing complete morphologies rather than primitives.
\end{document}